\documentclass{article}

    \PassOptionsToPackage{numbers, compress}{natbib}
    \usepackage[preprint]{neurips_2023}

\usepackage[utf8]{inputenc} % allow utf-8 input
\usepackage[T1]{fontenc}    % use 8-bit T1 fonts
\usepackage{hyperref}       % hyperlinks
\usepackage{url}            % simple URL typesetting
\usepackage{booktabs}       % professional-quality tables
\usepackage{amsfonts}       % blackboard math symbols
\usepackage{nicefrac}       % compact symbols for 1/2, etc.
\usepackage{microtype}      % microtypography
\usepackage{xcolor}         % colors
\usepackage{hyperref}
 
\usepackage{amsmath}
\usepackage[version=4]{mhchem} % chemical formulae
\usepackage{subcaption}     % 
\usepackage{graphicx}   

\title{Explainability Techniques for Chemical Language Models}

\author{%
  Stefan Hödl\\
  Physical Organic Chemistry, Radboud\\
  \texttt{stefan.hoedl@ru.nl} \\
  \And
  William Robinson\\
  Physical Organic Chemistry, Radboud\\
  \texttt{william.robinson@ru.nl}\\
  \And
  Yoram Bachrach\\
  Google DeepMind\\
  \texttt{yorambac@google.com}\\
  \AND
  Wilhelm Huck\\
  Physical Organic Chemistry, Radboud\\
  \texttt{w.huck@science.ru.nl}\\
  \And
  Tal Kachman \\
  Artificial Intelligence, Donders Institute, Radboud\\
  \texttt{talkachman@cerebrnita.com}\\
}

\begin{document}

\maketitle

\begin{abstract}
Explainability techniques are crucial in gaining insights into the reasons behind the predictions of deep learning models, which have not yet been applied to chemical language models. 
We propose an explainable AI technique that attributes the importance of individual atoms towards the predictions made by these models. 
Our method backpropagates the relevance information towards the chemical input string and visualizes the importance of individual atoms.
We focus on self-attention Transformers operating on molecular string representations and leverage a pretrained encoder for finetuning. 
We showcase the method by predicting and visualizing solubility in water and organic solvents. We achieve competitive model performance while obtaining interpretable predictions, which we use to inspect the pretrained model.
\end{abstract}

\let\clearpage\relax

\begin{figure}[!h]
\centering
  \begin{subfigure}[t]{\textwidth}
    \centering
    \includegraphics[width=\textwidth]{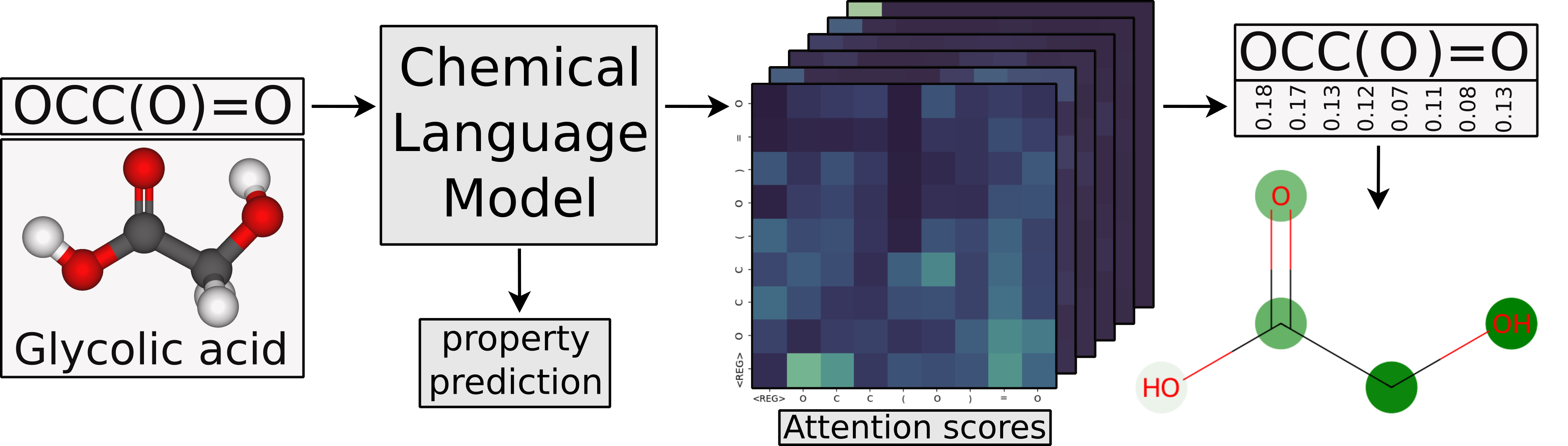}
    \label{fig:MolViz_left} 
  \end{subfigure}
  \caption{Schematic visualization of the explainability technique. The string representation ``OCC(O)=O'' of the molecule ``glycolic acid'' is the input to the chemical language model, which predicts the property of interest. The attention scores are extracted from all 6 Transformer layers (shown as stacked heatmaps). These attention scores are aggregated throughout the architecture following simple update rules to obtain a relevance weight for each token in the input string. The molecules' structure diagram is visualized with green highlights, where high color intensity corresponds to high attributed impact of an individual atom towards the predicted property.}
\label{fig:technique_schematic} 
\end{figure}

%%%%%%%%%%%%%%%%%%%%%%
\section{Introduction}
Transformers \citep{AIAYN, FormalTransformerAlgos} have achieved state-of-the-art results in vision \citep{ViT}, language~\cite{wolf2020transformers, otter2020survey} and life sciences \citep{AlphaFold2, ferruz2022controllable}. %
However, they are ``black boxes'' with millions of parameters: despite their excellent performance, they cannot provide explanations or human interpretable reasons behind their choices. 
This explainability gap severely limits their adoption and usage by practitioners, with models of high predictive accuracy sometimes breaking down in real world physical experiments \citep{Calibration_OOD_generalization,hogewind2022safe}. \\

Explainable AI (XAI) methods~\citep{dovsilovic2018explainable} are techniques that address this gap by visualizing, explaining or validating the predictions of machine learning models.
Visualizations of the importance of inputs or features to the model's predictions provide insights as to the reasons behind the predictions that the model makes, and can increase trust in the model's outputs, especially when these explanations correspond with human and scientific understanding of the task \citep{Perspective_MolViz}.\\

In the domain of chemistry, a set of molecules might have very similar properties despite diverse structural features, while the exchange of one functional group might dictate whether a molecule dissolves. 
Attributing the importance of individual atoms or functional groups towards the prediction can help chemists understand model predictions and leverage their intuition.
Various XAI techniques have been applied to molecules, frequently based on evaluating the importance by perturbation of the input, such as the model-agnostic SHAP method based on Shapley values~\citep{Perspective_MolViz, SHAP, NeuralPayoffMachines, VocabSelection_Shapley}.\\

Molecules are typically represented as molecular graphs or with subgraph-based ``fingerprints''~\citep{ECFPs, NeuralFPs},
significantly constraining the types of machine learning models and explainability techniques that can be applied. 
Specialized XAI techniques have been proposed \citep{GCNN_LRP, gCNN_XAI_mol} for graph neural networks (GNN)~\citep{GCN, geometricDL_5G} which represent molecules as molecular graphs, but these techniques may be ill-suited to explain models based on other representations. 
Molecules can also be represented as strings of letters and symbols, which are easily machine-readable but difficult to decode for humans.
One recent powerful tool for training chemical prediction models is applying Transformers to very large databases of such molecular strings, commonly known as ``chemical language models''~\citep{flam2022language,du2022molgensurvey}.
These models have shown impressive results on property prediction tasks \citep{Chemformer} and are especially promising when leveraged to generate novel molecules with specific properties \citep{RegressionTransformer, MolGPT_saliency, MolDreaming}.
However, explainability techniques for models based on molecular graphs or fingerprints are not applicable to chemical language models. \\

One technique specific to Transformer architectures is applying heatmap visualizations of the attention scores, typically extracted from the final attention layer in the architecture~\citep{MappingChemSpace_rawAttn}. This approach is limited as only the attention scores are considered, ignoring the previous layers' context and other components of the architecture.
Recent approaches overcome this limitation by considering all Transformer components and aggregating importance throughout the layers while retaining context \citep{AttnRolloutFlow, AttnViz_self}.
While attention-specific visualization techniques have been successfully used in computer vision tasks, they have not yet been adapted to the molecular domain.\\

\textbf{Our contribution.} Inspired by recent work on Transformer-based image classification ~\citep{AttnViz_self, AttnViz_MM}, we propose a framework for regression tasks operating on molecular strings, and develop an explainable AI technique for chemical language models, using solely the model without external tools or information. The attention visualization technique we propose for chemical language models allows identifying the individual atoms whose presence had the biggest impact on the output of a model predicting properties of the molecule.
We leverage a Transformer pretrained on molecular strings for finetuning on experimental chemical property measurements and achieve competitive predictive performance. This task-agnostic approach can easily be applied to all molecular regression and classification tasks.
In our approach, the relative impact that individual units have on the model's prediction are attributed to the input string and visualized.\\

We demonstrate the technique by predicting solubility of small molecules and assessing the model's predictions through visualizing attributed relevance.
Our demonstration is based on a well-studied task where large datasets are available, and where intuition can play a crucial role in validating the visualization explainability results. 
After visualizing relevance for aqueous solubility, we study solubility in organic solvents.
Our technique is directly applicable to regression and classification tasks using self-attention Transformer architectures on molecular string representations. It can be used following model training and only requires saving the attention scores and calculating their gradients.
Similarly, our method can be applied to other Transformer variants~\citep{FormalTransformerAlgos}, such as encoder-decoder architectures leveraging cross-attention, by applying modifications analogous to similar extensions in computer vision~\citep{AttnViz_MM}.
We fully \href{https://github.com/KachmanLab/Chemical\_Language\_Model\_Explainer}{open-source our methodology} for the scientific community.

%%%%%%%%%%%%%%%%%%%%%%
\section{Related work}
%%%%%%%%%%%%%%%%%%%%%%
\paragraph{Explainable AI.}
``Extrinsic'' XAI techniques are methods to visualize predictions of an already trained model, frequently based on attribution of the prediction to the model's inputs \citep{Perspective_MolViz}.
One example are salience maps, which produce local explanations of how changes in the input influence the output, measuring the squared norm of the model's gradient with respect to the input \citep{SaliencyMaps_CV}.
Many modern approaches are decomposition-based techniques leveraging gradients and propagation, such as layerwise relevance propagation \citep{Bach_LRP}, GradCAM \citep{GradCAM}, attention rollout \citep{AttnRolloutFlow}, gradient*input and DeepLIFT \citep{GradXInput, DeepLIFT_LRP}. 
All of these were initially proposed for computer vision classification tasks. \\

An extensive overview of current explainability techniques for molecules is provided by~\citet{Perspective_MolViz}. 
Due to their simplicity, saliency maps are commonly used in the chemical domain, for example to explain machine learning models based on molecular fingerprints~\citep{SaliencyMaps, gCNN_XAI_mol, MolGPT_saliency}.
Crippen's $\log(P)$ prediction is an additive approach which visualizes the contribution from each atom using experimental fits, a good baseline for heatmap approaches~\citep{CrippenLogP, do_machines_dream}. \\

However, many assumptions of these baseline XAI techniques, such as SHAP's additive features and fair coalitions, are not valid for chemical language models because they operate on molecular string representations.
Omission of some tokens frequently leads to invalid molecules, as these tokens directly represent the molecules' physical structure.
Determining valid substitution rules is difficult, and many valid substitutions would significantly change the predicted property.
SHAP can be applied to natural language, but its application to molecular string tokens is analogous to masking out random letters of a single word and evaluating importance of each letter.

\paragraph{Attribution of relevance.}
A related work is layerwise relevance propagation, a method designed for convolutional graph neural networks \citep{GCNN_LRP}.
The authors use a solubility regression task, but focus on adapting their approach to operate on molecular graphs and convolutional filters, where atoms and edges are represented explicitly and can thus be assigned relevance directly \citep{do_machines_dream}.\\

Theoretical justifications for layerwise relevance propagation are provided in the work of \citet{DeepTaylorDecomp}, which proposes the deep Taylor decomposition.
Heatmaps are defined as ``consistent'' if they fulfil conservation of relevance and yield only positive values without negative relevance.
The authors define a training-free relevance model for architectures using ReLU nonlinearities which have positive activations.
This allows them to propagate relevancies in higher layers in proportion to the model's activations and decompose the architecture layer by layer to the input \citep{DeepTaylorDecomp}.
% \[ R_i = \sum_j \frac{z_{ij}^+}{ \sum_{i`} z_{i'j}^+ } R_j\]
%based on their $z^+$ rule 

\paragraph{Molecular string representations.}
There are many ways to represent a molecule as an input to a machine learning model, which has a high impact on the predictive performance and ease of use by chemists~\citep{wigh2022review}. 
One prominent such language is {SMILES} \citep{SMILES}, a ubiquitous machine-readable format representing molecules as strings, which consists of letters, symbols, digits and brackets, to denote atoms, bonds, rings and branches. 
SMILES has been in use since the 1980s, and offers a language that is flexible yet easy to understand and communicate~\citep{weininger1988smiles}. 
A SMILES representation can be derived algorithmically through a depth-first graph traversal and iterative breaking and labeling of cycles and branches.
Multiple valid SMILES strings can be obtained through different numbering choices, thus a ``canonical'' version was developed to uniquely identify molecules.
However, invalid SMILES strings can easily be constructed, which is addressed by the SELFIES~\citep{SELFIES_advances} notation frequently applied in generative settings.
Transformers can take a SMILES string directly as input after the tokenizer splits the string into its constituent tokens.

\newpage %arxiv
\paragraph{Solubility prediction.}
We demonstrate and analyze our approach using the regression task of predicting the solubility of small molecules, a difficult but well-studied problem crucial to many applications~\citep{lusci2013deep}.
Many approaches have been proposed to predict solubility, ranging from simple additive contributions using empirical measurements to elaborate deep learning architectures.
The property of interest is the log-scaled solubility $\log(S)$, the solid-solution partitioning coefficient, in units of [mol/L]~\citep{atkins2014atkins}.
While datasets of $\log(S)$ calculations are available, a crucial limiting factor is the availability of experimental solubility measurement datasets. Aqueous solubility datasets are aggregated, such as ESOL ($\sim$3K) \citep{ESOL} and AqSolDB ($\sim$10K) \citep{AqSolDB}. % contain 3K-10K measurements of aqueous solubility at room temperature.
SolProp is the largest available solubility data collection, which aggregates experimental measurements for solubility in water (\textit{AqueuosSolu}, $\sim$12K) as well as solid solubility in organic solvents (\textit{CombiSolu-Exp}, $\sim$5K)~\citep{SolProp}.\\ %\newpage

The SolProp authors further propose their own predictive approaches based on GNNs to predict solubility in water at room temperature $\log(S_{\mathrm{aq, 298K}})$.
For solubility in an organic solvent $X$ at temperature $T$, GNN predictions of the free energy $\Delta G_{\mathrm{solv}}$ and enthalpy $\Delta H_{\mathrm{solv}}$ of the thermodynamic cycle as well as the $\log(S_{\mathrm{aq, 298K}})$ prediction are combined to infer $\log(S_{X, T})$.
All GNN models are pretrained on computational datasets of $\Delta G_{\mathrm{solv}}$, $\Delta H_{\mathrm{solv}}$ where available and finetuned on experimental datasets.
Particularly interesting is the ability of their approach to incorporate available experimental measurements to drastically improve the predictive performance \citep{SolProp}.

\paragraph{Transformers for solubility.}
Transformer-based architectures have been explored to directly predict aqueous solubility, among them SolTranNet \citep{SolTranNet} which adapts the MoleculeAttentionTransformer \citep{MoleculeAttnTransformer} architecture for the AqSolDB dataset.\\

Another recent work pretrains a SMILES language model from scratch, focusing on directly predicting the solvation free energy as well as solubility in organic solvents.~\citep{SolvBERT}
Similar to SolProp, most of their models are pretrained on the computational datasets, but only the structures and not the calculated properties are used during self-supervised pretraining via masking. Models trained on multiple combinations of pretraining and finetuning datasets are presented, but no results for the solubility datasets of the SolProp collection are available for comparison (\textit{AqueousSolu}, \textit{CombiSolu-Exp}).

%%%%%%%%%%%%%%%%%%%%%%
\section{Methods}
%%%%%%%%%%%%%%%%%%%%%%
\paragraph{Chemical language model.}
As the investigated Transformer model, we use the encoder of MegaMolBART~\citep{MegaMolBART, MolBERT}, a self-attention architecture based on the Chemformer \citep{Chemformer}. 
Formal algorithms for Transformers are presented in \citep{FormalTransformerAlgos}. We consider an encoder Transformer, and more specifically a bidirectional self-attention Transformer. % where masks are only applied as padding. 
Within an attention layer, $\mathbf{Q}$ and $\mathbf{K}$ are the query and key projections, their element-wise dot product is scaled by $\sqrt{d_k}$ for numerical stability. 
The attention scores $\mathbf{A}$ are calculated as
\begin{equation} \label{eq0:attn}
\mathbf{A} = \text{softmax}\left(\frac{\mathbf{Q}\mathbf{K}^T}{\sqrt{d}}\right).
\end{equation}

The Chemformer is pretrained in a self-supervised fashion (BART \citep{BART}) using a reconstruction loss, which is obtained by masking random parts of the input tokens with the <MASK> token.
The model thus learns to reconstruct the original SMILES string, which requires only strings but no labels as input and is crucial to achieve scale. 
The primary difference between the Chemformer and MegaMolBART is scale, where MegaMolBART was pretrained on approximately 1.45 billion small molecules from the ZINC-15 database \citep{ZINC15} while keeping the size of the model constant.\\

Only the encoder is used for regression or classification tasks, which can be viewed as leveraging the pretrained model as a black box to extract expressive latent features. 
A small hierarchical regression head with dense layers, nonlinearities and layer normalization takes the average of these feature vectors from the encoder and transforms it into a scalar prediction of the property of interest.
The model is subsequently finetuned on the prediction task of interest using a dataset of experimental measurements (\textit{transfer learning}).
This enables the model to rapidly converge and achieve competitive accuracy even for small datasets using very few computational resources compared to training from scratch. 
In principle, only the $\sim$40K parameters of the regression head need to be trained, while the $\sim$20M parameters of the pretrained encoder with 6 layers can be frozen. 

\paragraph{<REG> tokenization.}
A tokenizer splits the molecules' SMILES string into a list of tokens of variable length using regular expressions. 
These tokens are the input to the encoder, which outputs a latent representation for every token. 
A pooling step is required to obtain a feature vector of static shape for the regression head, usually the average or sum of all tokens' latent representations is used.
However, this pooling step hinders attribution of importance through backpropagation of the prediction to the input string, since the contribution of each token is lost in the average or sum.
Due to this pooling, saliency maps of the encoders' output show uniform relevance for all tokens.

We avoid this pooling step by prepending a <REG> token to the tokenized SMILES string, which aggregates information throughout the encoder.
Only the feature vector of this <REG> token is processed in the regression head, which allows us to obtain the relevance of all other token to the <REG> token from the pairwise attention weights and backpropagate gradients from the regression prediction to the input tokens.
The vocabulary of the MegaMolBART tokenizer thus needs to be extended with a <REG> token corresponding to the <CLS> token frequently encountered in the domain of computer vision and natural language processing for classification tasks~\citep{BERT, BART, ViT}.

\paragraph{Relevance aggregation.}
Following \citet{AttnViz_MM}, for self-attention Transformers the relevancy matrix $\mathbf{R}$ is initialized as the identity matrix $\mathbf{R}^{0} = I^{t \times t}$ for a molecule with $t$ tokens, representing self-contained relevance of each token. 
The attention scores $\mathbf{A}^l$ of each Transformer layer $l \in L$ are saved during the models' forward call, while the prediction's gradients $\nabla \mathbf{A}^l := \frac{\partial y}{\partial A^l}$ are evaluated and saved during the backward call.
In multi-head attention, each attention head captures different aspects of the task, the importance of each attention head can be quantified from its gradient $\nabla \mathbf{A}$. 
To account for these differences, the attention scores $\mathbf{A}$ are multiplied by their gradient before averaging over all 8 attention heads ($\mathbb{E}_h$).
Only the positive importance $(\cdot)^{+}$ is considered to reflect nonlinearities following the deep Tayler decomposition theory, $\odot$ denotes the element-wise product.
\begin{equation} \label{eq1:abar}
\bar{\mathbf{A}}^{l} = \mathbb{E}_h \big( (\nabla \mathbf{A}^{l} \odot \mathbf{A}^{l})^{+} \big)
\end{equation}

The previous layers' relevance matrix $\mathbf{R}^{l-1}$ contextualizes the current layers attention mechanism, thus it is matrix multiplied with the layer's contribution $\bar{\mathbf{A}}^l$ and added element-wise to $\mathbf{R}^{l-1}$ reflecting skip connections.
Relevance is aggregated throughout all layers $l \in \{1, 2, \dots, 6\}$ with the update rule 
\begin{equation} \label{eq2:aggregation}
\mathbf{R}^{l}= \mathbf{R}^{l-1} + \bar{\mathbf{A}}^{l} \cdot \mathbf{R}^{l-1}.
\end{equation}

The feature vector of the <REG> token encapsulates all information contributing to the prediction, and the relevance of each token to the <REG> token is captured in the corresponding entry in the contextualized relevancy matrix $\mathbf{R}$. 
As a consequence, visualization of the final layers' averaged attention matrix $\bar{\mathbf{A}}^6$ shows zeros everywhere but in the this column.

\begin{figure}[!h]
\centering
  \begin{subfigure}[t]{0.40\textwidth}
    \centering
    \includegraphics[width=0.99\textwidth]{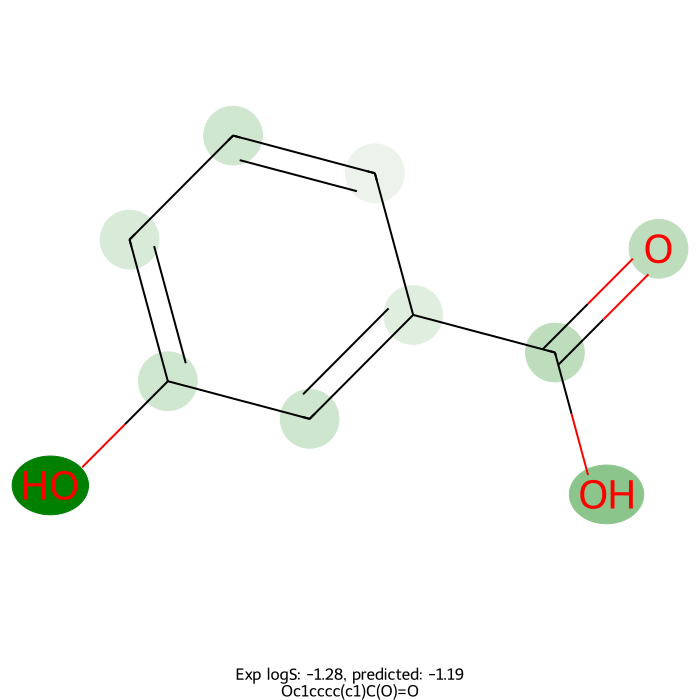}
  \end{subfigure}
  \hfill
  \begin{subfigure}[t]{0.43\textwidth} 
    \centering
    \includegraphics[width=0.99\textwidth]{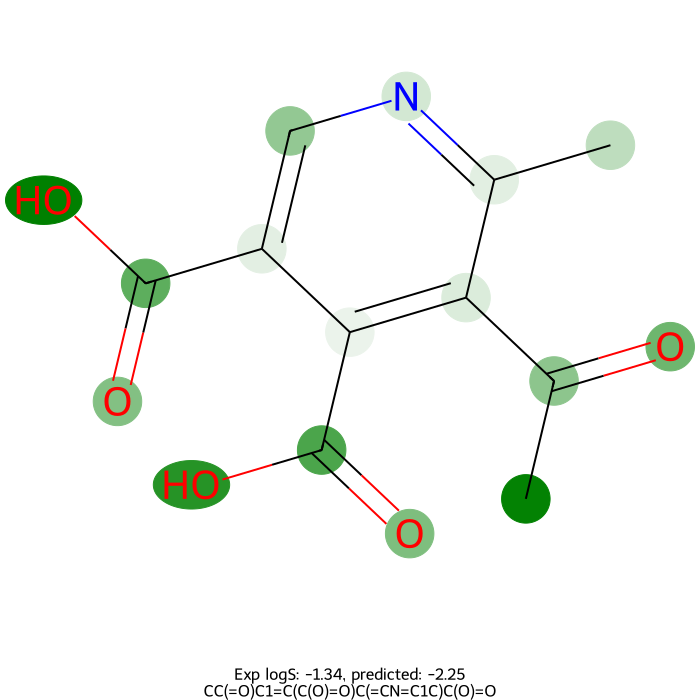}
  \end{subfigure}
  \hfill
  \begin{subfigure}[t]{0.15\textwidth}
    \centering
    \includegraphics[width=0.6\textwidth]{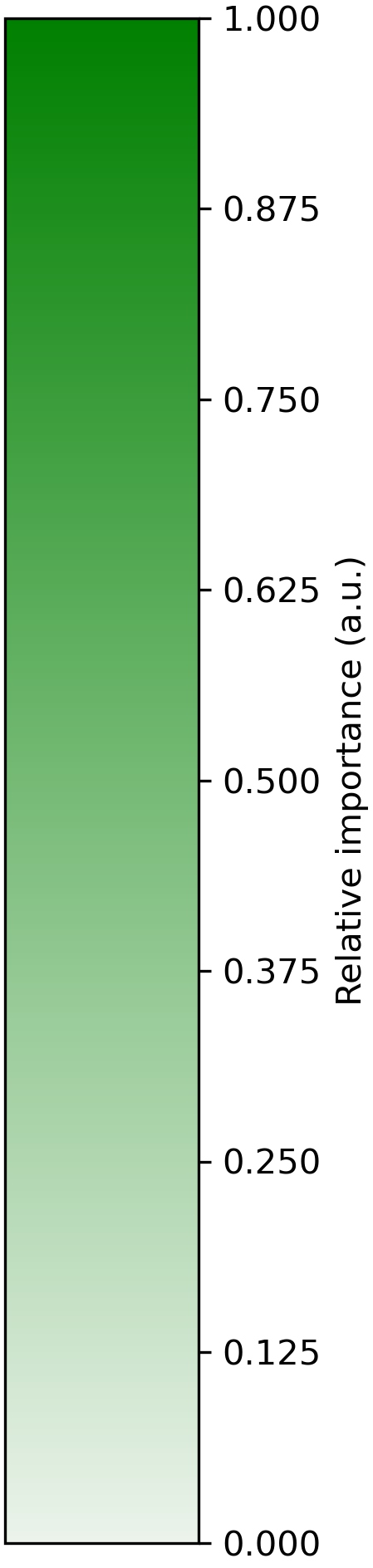}
  \end{subfigure}
\caption{2d structure plot with a color map corresponding to attributed relevance extracted from the <REG> token of the aggregated relevancy matrix $\mathbf{R}$. The relevance of each atom token towards the model's aqueous solubility prediction is visualized, the intensity of the green highlights corresponds to the attributed relevance. The color bar shows relative importance in arbitrary units.
}
\label{fig:MolViz} 
\end{figure}

%arxiv: swapped paragraph + fig
\paragraph{Relevance visualization.}
All atom token relevance weights are then visualized on the molecules' 2D structure diagram to aid human intuition, as understanding SMILES strings intuitively is difficult even for experienced practitioners. 
Once the attributed importance weights are obtained, the molecule is visualized by plotting a 2d structure diagram with a corresponding heatmap showing the attributed relevance to the prediction (Fig. \ref{fig:MolViz}).
% \newpage %arxiv
% \newpage
%%%%%%%%%%%%%%%%%%%%%
\section{Experiments}
%%%%%%%%%%%%%%%%%%%%%%
We first present prediction errors from the finetuned MegaMolBART encoder model for aqueous solubility and compare them to results from other approaches available in the literature. We then turn to examining solubility in organic solvents.
After establishing the models' accuracy, we obtain the internal representations to visualize and inspect the attributed importance for a range of molecules. 

\paragraph{Aqueous solubility prediction.}
The results below show that when predicting aqueous solubility $\log(S_{\mathrm{aq, 298K}})$, the predictive accuracy of the finetuned MegaMolBART model is competitive, but slightly worse than the corresponding SolProp GNN model.
The SolProp authors evaluate their model on a randomly selected as well as a low-uncertainty test set of the AqueousSolu dataset, as significant uncertainty in experimental measurements is introduced due to aggregation from diverse literature sources and multiple datasets.
The model errors of the finetuned MegaMolBART model fall behind the specialized SolProp GNN model on both the random and the accurate test set.
The SolProp authors also evaluate ALOGpS \citep{ALOGPS_VCCLab} and SolTranNet \citep{SolTranNet} on the accurate test set (Table \ref{table:aqueousSolu}). 

\begin{figure}[!h]
\centering 
  \begin{subfigure}[t]{0.47\textwidth}
    \centering
    \includegraphics[width=0.99\textwidth]{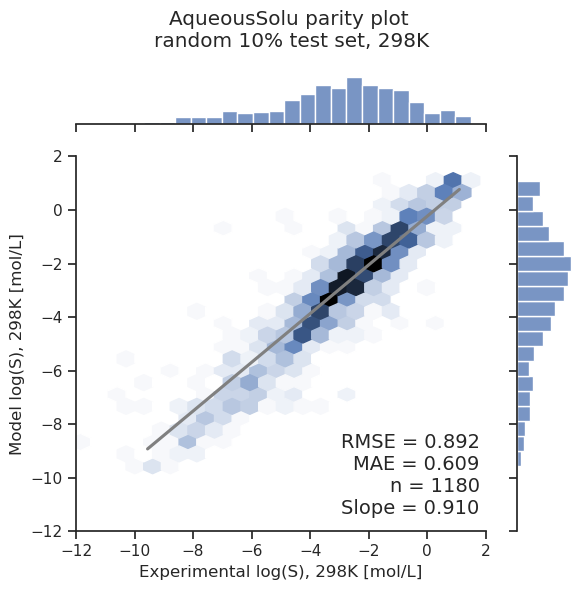}
  \end{subfigure}
  \hfill
  \begin{subfigure}[t]{0.47\textwidth}
    \centering
    \includegraphics[width=0.99\textwidth]{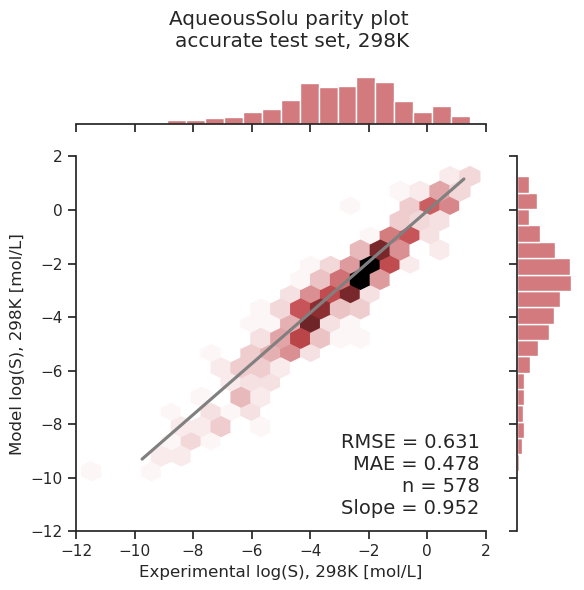}
  \end{subfigure}
\label{fig:parityPlots} 
\caption{Parity plots showing the predictive accuracy of the MegaMolBART model finetuned on the AqueousSolu dataset. The left plot (blue) shows a (10\%) randomly selected test set, while the right plot (red) shows the low-uncertainty test set \citep{SolProp}.
The experimental solubility measurements $\log(S_{\mathrm{aq, 298K}})$ (X axes) are plotted against the models' predictions (Y axes), deviations from the diagonal represent the models' errors. The predictions are grouped into hexagonal bins to highlight the density of predictions, the raw densities are shown as histograms on the sides of both plots. The grey lines show a linear regression fit and its ``slope'' with $n$ samples in the test set. The run closest to the average of 5 runs with different seeds for the dataset split is shown.}
\end{figure}

\begin{table}[!h]
  \caption*{Model comparison for aqueous solubility}
  \centering
  \begin{tabular}{lcccc}
    \toprule
    Model        & Random & Random & Accurate & Accurate \\
                 & MAE      & RMSE      & MAE        & RMSE   \\
    \midrule    
    MegaMolBART (AVG)  & 0.604 & 0.895    & 0.475   & 0.638  \\
    MegaMolBART (SE) & (0.004) & (0.009)    &  (0.006)   & (0.006)  \\
    \cmidrule(r){2-3} \cmidrule(r){4-5}
    SolProp      & \textbf{0.49} & \textbf{0.75} & \textbf{0.34} & \textbf{0.49} \\
    ALOGpS       & -    & -    & 0.55 & 0.79 \\
    SolTranNet   & -    & -    & 0.58 & 0.76 \\
    \bottomrule
  \end{tabular}
  \caption[AqueousSolu prediction error comparison]{Comparison of models trained on the AqueousSolu dataset for random 10\% test sets (``Random'') and the low-uncertainty test set (``Accurate''). 
  The mean (AVG) and standard error (SE) of the mean absolute error (MAE) and root mean squared error (RMSE) are reported, calculated from 5 runs with random seeds affecting the split into both train and validation sets and the random test set.
  }
  \label{table:aqueousSolu}
\end{table}

\begin{figure}[!h]
\centering
  \begin{subfigure}[t]{\textwidth}
    \centering
    \includegraphics[width=0.99\textwidth]{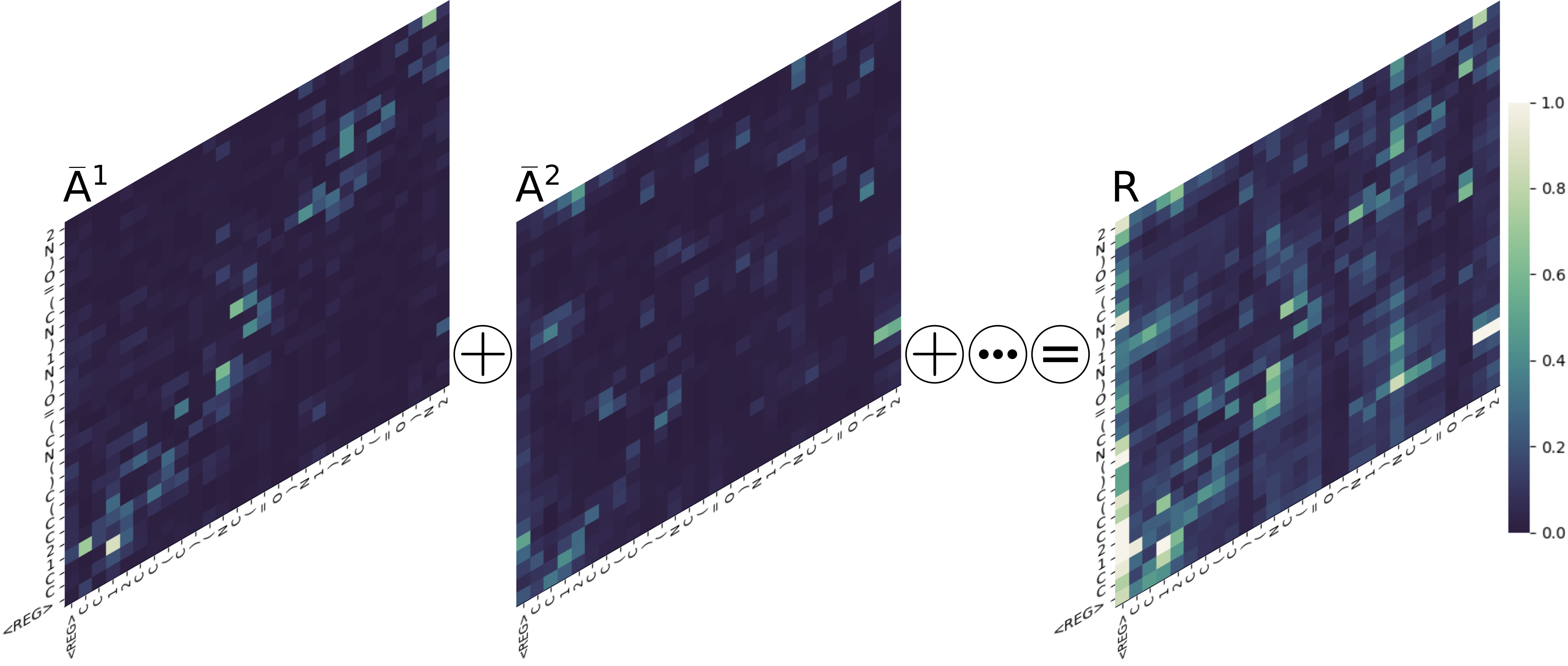}
  \end{subfigure}
\caption{Visualization of the attention aggregation (Eq. \ref{eq1:abar} and \ref{eq2:aggregation}). The $\bar{\mathbf{A}}$ heatmap is aggregated throughout all 6 attention layers, of which layer 1, 2 and the aggregated relevancy matrix $\mathbf{R}$ are shown for the molecule ``CC12CC(C)(NC(=O)N1)NC(=O)N2''. The relevancy matrix $\mathbf{R}$ is obtained by starting with the identity matrix $I$ and applying the update rules using the averaged attention matrix $\bar{\mathbf{A}}$ extracted from all 6 attention layers.}
\label{fig:stackedAttnHeatmap} 
\end{figure}

\paragraph{Attention heatmap visualization.}
Fig. \ref{fig:stackedAttnHeatmap} shows heatmaps of the first two layers' averaged attention scores $\bar{\mathbf{A}}$ to illustrate the aggregation rules. 
The pairwise attention between each token can be seen, where light colors reflect large attention scores.
Regions of high importance can be seen near the diagonal of the first heatmap $\bar{\mathbf{A}}^1$, which captures the immediate neighbourhood of each atom.
Brackets and digits denote side-branches and rings, which can lead to close physical proximity of two atoms despite large separation in the SMILES string.
High attention scores can be seen for ring openings and closures, in particular for the carbon atoms in close physical proximity.
The final aggregated heatmap $\mathbf{R}$ is plotted on the right side after removal of the identity matrix $I$ to account for differences in scale. 
The <REG> token's column of $\mathbf{R}$ is extracted and visualized.

\paragraph{Solubility in organic solvents.}
For solubility prediction in different solvents, the interactions between the solute and the solvent need to be taken into account in the architecture.
The MegaMolBART model is finetuned on the CombiSolu-Exp dataset in the same fashion as for aqueous solubility. The visualization technique can be applied without further modifications as long as the task can be formulated within an architecture based only on self-attention, which is achieved through the usage of the <SEP> token to separate the SMILES string of the solute and solvent.
This simplifies the task as well as the visualizations of the solute and solvent, as the model receives a single concatenated input to process, and thus only a single relevancy matrix $\mathbf{R}$ needs to be constructed.
The tokenized input for glycolic acid ( OCC(O)=O ) shown in Fig. \ref{fig:technique_schematic} in ethanol ( CCO ) as a solvent would then be %\vspace{0.5ex}

\makebox[\textwidth]{
`` <REG>, O, C, C, (, O, ), =, O, <SEP>, C, C, O ''. 
}%\vspace{-0.1ex}

\paragraph{Cross-attention model variant.}
An alternative approach is to model the interaction using cross-attention layers, where the query and key vectors correspond to the latent representations of the solute and solvent.
Both are extracted from the MegaMolBART's encoder and averaged into two vectors of static shape independent of the number of tokens in the solute and solvent.
One or more cross-attention layers are used to model the interaction between solute and solvent, which then needs to be aggregated into a single feature vector for the regression head.
This is achieved through the <REG> token in the aforementioned approach but does not work here due to the pooling operation before the cross-attention layers. 
It contradicts the technique's assumption that the <REG> token in the aggregated relevance matrix $\mathbf{R}$ reflects the relevance of each other token to the predicted property, which are no longer explicitly represented per token after pooling.
Separate attention scores and gradients need to be stored to obtain relevancy matrices $\mathbf{R}_\mathrm{solute}$ and $\mathbf{R}_\mathrm{solvent}$, which requires extensions to the propagation rules following \citet{AttnViz_MM}.
Separate heatmap visualizations can be obtained from the cross-attention layers, which might explicitly highlight interactions. \\ %\newpage
%[TODO: do cross-attention for solute, solv, THEN concatenate with <REG> [solu] <SEP> [solv]]

The predictive accuracy of both models is excellent, though a direct comparison cannot be made due a big difference in approach and datasets used to model solubility in organic solvents.
The cross-attention architecture variant achieves very accurate results but cannot be explained through the <REG> token approach without further modifications to the explainability technique.
Since the SolProp authors use GNN predictions of $\Delta G_{\mathrm{solv}}$, $\Delta H_{\mathrm{solv}}$, $\log(S_{\mathrm{aq, 298K}})$ and thermodynamic cycle relationships to predict $\log(S_{X, T})$, their model is not trained on the CombiSolu-Exp dataset itself. 
However, it does use the GNN models trained on AqueousSolu and corresponding datasets of $\Delta H_{\mathrm{solv}}$ and $\Delta G_{\mathrm{solv}}$. We refer to the SolProp paper and supplementary information for results in various contexts~\citep{SolProp}.
This implies that visualization techniques would need to consider all SolProp architecture components.
In contrast, both MegaMolBART models are trained exclusively on the CombiSolu-Exp dataset. %, which prevents direct comparisons. 

\begin{figure}[!h]
\centering
  \begin{subfigure}[t]{0.42\textwidth}
    \centering
    \includegraphics[width=\textwidth]{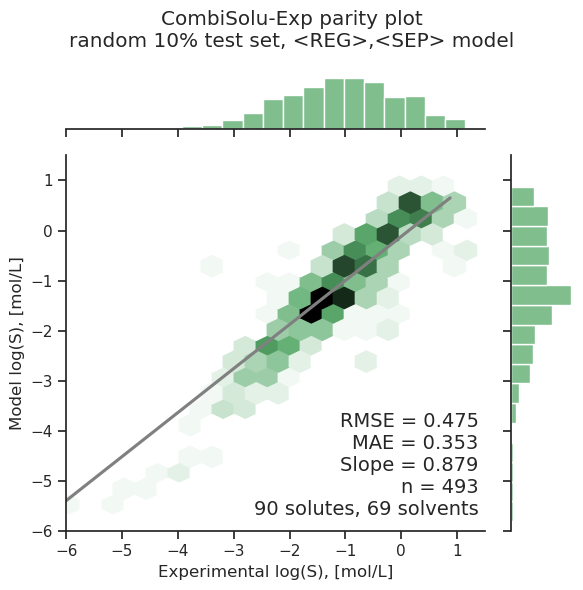}
  \end{subfigure}
  \hfill
  \begin{subfigure}[t]{0.42\textwidth}
    \centering
    \includegraphics[width=\textwidth]{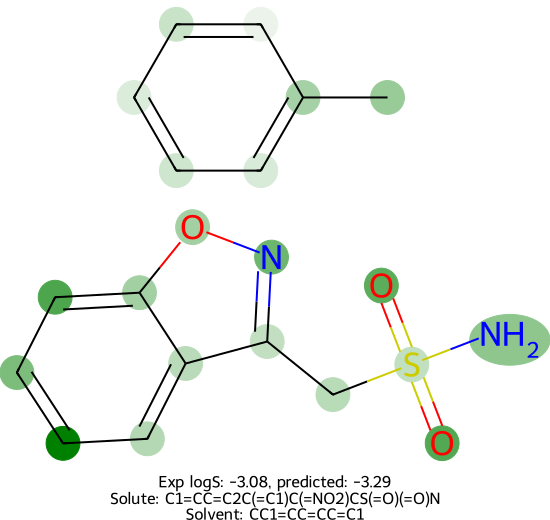}
  \end{subfigure}
  \hfill
  \begin{subfigure}[t]{0.14\textwidth} 
    \centering
    \includegraphics[width=0.55\textwidth]{figs/colorbar_AU.png}
  \end{subfigure}
\caption{
The left parity plot (green) shows the finetuned MegaMolBART models' predictive accuracy for solubility in organic solvents using the <SEP> token to concatenate the solute and solvent tokens. 
It is evaluated on 5 randomly selected 10\% test sets of the CombiSolu-Exp dataset, the best run's parity plot is shown. The average and standard error as well as the corresponding parity plot for the cross-attention model evaluated on the $\mathrm{T} = \mathrm{298K}$ test set from \citep{SolProp} can be found in the supplementary information. The right visualization shows the attributed relevance and corresponding color bar of the <REG> model for a solute in an oil (toluene) as a solvent.
}
\label{fig:CombiSolvParity_MolViz} 
\end{figure}

\paragraph{Solute $\times$ solvent visualization.}
The relevancy matrix from the approach leveraging the <REG> and <SEP> tokens is obtained in the same manner as for aqueous solubility by aggregating relevancy throughout the layers for the concatenated input. 
The importance of the solute and solvent towards the solubility prediction is visualized in Fig. \ref{fig:CombiSolvParity_MolViz}. 
Different functional groups of a molecule contribute towards solubility when the same solute is dissolved in distinct solvents, such as an oil (toluene) and an alcohol (ethanol).
However, when the models' $\log(S)$ predictions are close to each other, the importance attributed to both solutes is nearly indistinguishable despite distinct interactions contributing to solubility. %
These interactions are not captured by the <REG> model and might be the reason why the cross-attention model variant is able to outperform the <REG> model on the $\mathrm{T} = \mathrm{298K}$ test set of the CombiSolu-Exp dataset.

\paragraph{Discussion of results.}
We worked with experimental organic chemists to verify our methodology by showing them visualizations of the models' predictions to understand the nature and quality of the finetuned model. 
Our observations suggest the MegaMolBART model learns to predict solubility accurately without fully learning the relevant physical context.
Atoms with high attributed relevance generally correspond to relevant functional groups for aqueous solubility (such as \ce{OH}, \ce{=O} or \ce{NH2}), while explanations deteriorate for molecules with very high or low solubility. 
\newpage
The model is unable to accurately model symmetry and frequently attributes very different relevance  to symmetric functional groups. 
Interestingly, the model frequently attributes high relevance to the molecules' carbon backbone, which dictates the structure of the molecule but does not by itself influence solubility.
A possible explanation is that the model learns to predict solubility using the carbon backbone instead of functional groups due to the pretraining procedure of MegaMolBART.\\ %\newpage

Our results indicate that different finetuning approaches and regression head variants produce quantitatively comparable results and qualitatively similar visualizations due to the major influence of the pretrained MegaMolBART model. Similarly, the <REG> approach and the average-pooling achieve very similar accuracy on the AqueousSolu dataset, thus the <REG> token enables attribution of relevance and visualizations without any loss in accuracy.

% \newpage %temporary page break for consistent formatting with figures
%%%%%%%%%%%%%%%%%%%%%%
\section{Limitations \& future work}
%%%%%%%%%%%%%%%%%%%%%%
The presented explainability technique could be significantly improved by visualizing relevance attributed to all tokens in a molecular string, instead of only atom tokens. SMILES strings contain tokens which do not directly correspond to atoms or bonds, and thus do not have an explicit correspondence to structure diagrams.
This can be achieved by constructing a mapping between non-atom SMILES tokens and 2d structure plots, mapping relevance of non-atomic tokens such as ring opening and closures, brackets and bonds. Crucially, no ground truth relevancy labels exist, which makes validation of one approach over alternatives difficult.\\

A naive approach would be to assign relevance attributed to bond tokens to the bonding pair and distribute bracket and ring token relevancies to all constituent atoms. 
Another possible approach to address this issue is to initialize the relevancy matrix as 1 only for atom tokens, but this prevents the propagation of relevance between tokens throughout the layers.
Alternatively, a sum over additive contributions for all atom tokens could be extracted instead of the <REG> token, but this suffers from the same explainability issues as with the pooling layer.
While naively disregarding all non-atom tokens biases the visualization of the attributed importance, different variations of the approach similarly introduce biases and assumptions.
Plotting all explicitly represented bonds does not yield a more accurate importance either but requires adapting SMILES strings and training from scratch.\\

Visualizations as well as predictions of molecular properties should also be symmetric and invariant to different representations of the same molecule.
Non-canonical SMILES string can be obtained by different choices of labeling of a given molecule.
The MegaMolBART model does not fulfil invariance to randomized SMILES inputs even though SMILES randomization was used as data augmentation during training \citep{MegaMolBART}.
Contrastive training approaches have been proposed \citep{FragNet_contrastiveSMI} to enforce SMILES invariance in the model, but this is outside of the scope of this study due to computational efforts required to pretraining the model from scratch.\\

Finally, the method is applicable to Transformer architectures beyond self-attention, where generative architectures in particular would greatly benefit from visualization techniques.  However, further work and additional modifications are necessary to apply the explainability technique, among them the extraction and propagation of multiple relevancy matrices.

%%%%%%%%%%%%%%%%%%%%%%
\section{Conclusion}
%%%%%%%%%%%%%%%%%%%%%%
With the advent of large language models in chemistry, increasing trust in their predictions is a crucial step towards incorporating deep learning model into the laboratory. 
Visualization of relevance of molecules' atoms and functional groups enables practitioners to inspect the learned features of chemical language models and validate their correspondence to scientific understanding and intuition. \\

We propose an attribution technique for Transformers operating on molecular string representation, which addresses the lack of attention-specific explainability techniques for chemical language models.
Our method propagates importance throughout the models' layers and attributes these to the input SMILES string components using only the attention scores and their gradients. The <REG> token encapsulates the relevance of each token towards the property prediction, which is extracted and visualized. \\

We demonstrate the effectiveness of our technique by predicting and visualizing solubility in water and organic solvents using a task-agnostic approach, achieving competitive accuracy and explainable predictions. These visualizations allow us to inspect what the chemical language model learns.
Our technique is applicable to any self-attention Transformer architectures and molecular string representations, and can be extended to other Transformer architectures and tasks. %and can be extended to encoder-decoder architectures.

%\newpage

% \section*{References}
\bibliography{references}

\newpage

\section{Supplementary Information}

\paragraph{Hyperparameters.}
Relatively few hyperparameters need to be selected due to the pretrained MegaMolBART model. Most variations of the regression head produce very similar results, thus a relatively small regression head of dimensionality 512 - 64 - 64 - 1 is chosen. This aligns with the MegaMolBART output of dimensionality 512 (8 heads of dimensionality 64), and produces similar accuracy to bigger variants. Interleaved are ReLU nonlinearities as well as a LayerNorm layer before the first projection.
The regression head has 38.1K parameters, of which the first dense layer (512 $\rightarrow$ 64) has 32.8K, the second (64 $\rightarrow$ 64) 4.2K, and the final (64 $\rightarrow$ 1) 65 parameters. The LayerNorm adds another 1024 parameters. For comparison, the MegaMolBART encoder has 19.8M parameters. 
For the CombiSolu-Exp dataset, the scalar temperature is appended to the <REG> token feature vector for the regression head. Solvent density is not used as is detrimental to the model's accuracy.\\

The HuberLoss loss criterion is chosen to balance the mean absolute error (MAE) and root mean squared error (RMSE). The AdamW optimizer with betas 0.9 and 0.999 is used, the learning rate is dynamically adjusted by pytorch lightning. Finetuning only the regression head produces comparable results to finetuning all parameters. A finetuning-scheduler to incrementally unfreeze an additional layer is also explored but produces similar results while introducing training issues related to extraction of the attention scores and gradient hook. SMILES randomization as data augmentation during finetuning is explored but consistently yields significantly worse results.

\paragraph{Datasets.}
Two datasets are used, respectively the AqueousSolu dataset for solubility in water, and the CombiSolu-Exp dataset for solubility in organic solvents. Both are from the SolProp data collection (https://zenodo.org/record/5970538). For AqueousSolu the ``accurate'' setting proposed by the SolProp authors is obtained by selecting all molecules with more than 1 measurement, where the standard deviation does not exceed 0.2. 
This yields 578 test set molecules, the source of the mismatch of 1 compared to SolProp (579) is unknown. One extreme outlier with $\log(S)$ = 6.4 is also removed from the dataset. \\

For the CombiSolu-Exp dataset, the ``$\mathrm{T} = \mathrm{298K}$'' test set is obtained by selecting all molecules with a temperature which rounds to 298K. Samples which contain missing ``temperature, solvent density or experimental $\log(S)$'' values are excluded. The $\log(S)$ measurements of the CombiSolu-Exp dataset are scaled to the range [0, 1] for training stability.
For the ``random'' test setting in both datasets, a 10\% test set obtained from a random seed is set aside. The remaining molecules are split into a 90\% training and a 10\% validation set. The random seed only affects the split into train and validation set in the ``accurate'' and ``$\mathrm{T} = \mathrm{298K}$'' setting.

\paragraph{MegaMolBART.}
The pretrained MegaMolBART model is obtained through the provided docker container image (version v0.2:0.2.0) from github.com/NVIDIA/MegaMolBART. 
The pretrained pytorch model and weights are accessed directly to avoid the latency of the InferenceWrapper. 
The RegExTokenizer vocabulary is extended with a <REG> token, the tokenize function similarly needs to be adjusted. 
The encode function of MegaMolBART takes the SMILES string and produces latent representations of dimensionality (n\_tokens, 512), which includes <PAD> tokens up to pad up to the maximum token size of the batch. 

\paragraph{Extracting attention scores \& gradients.}
To extract the attention scores and gradients, the transformers.py file needs to be extended with a few lines of code to save and retrieve the attention scores and attach a hook to evaluate the gradients during inference. These hooks do not need to present during training but can also be extended afterwards. It should be noted that the validation, test and predict functions require enabling the gradient (torch.set\_grad\_enabled(True)) to obtain the gradients of the attention scores due to the backwards hook, which is enabled when training = False.

\paragraph{Computation cost.}
All models were finetuned on a single Nvidia RTX 2080 GPU, which allows for a maximum batch size of $\sim48$ due to memory constraints. A full finetuning run with 50 epochs takes $\sim10$ minutes for AqueousSolu and $\sim3$ minutes for 40 training epochs on CombiSolu-Exp.
Obtaining the attention scores and gradients for the explainability technique is very computationally efficient, as only a single forward and backward pass of the model is necessary, while the propagation and attribution of relevance is negligible in terms of computational cost.

% \newpage
\subsection{Model results.}
In addition to the main AqueousSolu results (Table 1), the following table presents errors from 5 runs keeping the data seed for the split into train, validation and test set identical while randomizing the model seed which affects the initialization of the regression head. Thereafter, results for the ``$\mathrm{T} = \mathrm{298K}$'' test set of the CombiSolu-Exp dataset are presented.

\begin{table}[!h]
  \caption*{AqueousSolu results with random model seed}
  \centering
  \begin{tabular}{lcccc}
    \toprule
    % \multicolumn{2}{c}{Part}                   \\

    Model        & Random       & Random       \\%& Accurate      & Accurate \\
                 & MAE (SE)     & RMSE (SE)    \\%& MAE (SE)      & RMSE (SE)  \\
    \midrule    
    Average  & 0.602  & 0.867    \\%   & -             & - \\
    Standard Error & (0.002) & (0.004)   \\%
    \bottomrule
    % \midrule    
  \end{tabular}
  \caption{Model error of MegaMolBART trained on the AqueousSolu dataset using 5 runs with random model seeds affecting the initialization of the regression head.}
  \label{table:aqueousSolu_modelSeed}
\end{table}

\begin{table}[!h]
  \caption*{CombiSolu-Exp results with random data seed}
  \centering
  \begin{tabular}{lcccc}
    \toprule
    % \multicolumn{2}{c}{Part}                   \\

    Model        & Random       & Random        \\
                 & MAE (SE)     & RMSE (SE)     \\
    \midrule    
    Average  & 0.378 & 0.558    \\
    Standard Error & (0.012) & (0.024) \\
    % \cmidrule(r){2-3} \cmidrule(r){4-5}
    \bottomrule
  \end{tabular}
  \caption{Model error of MegaMolBART trained on the ``random'' test set of the CombiSolu-Exp dataset using 5 runs with random data seeds affecting the split into train, validation and test set.
  %The mean (AVG) and standard error (SE) of the mean absolute error (MAE) and root mean squared error (RMSE) are reported.
  }
  \label{table:aqueousSolu}
\end{table}

For the CombiSolu-Exp ``$\mathrm{T} = \mathrm{298K}$'' test set, the <REG>,<SEP> model variant performs noticeably worse. 
An alternative approach is to model the interaction using cross-attention layers, where the query and key vectors correspond to the averaged latent representations of the solute and solvent, which achieves state-of-the-art accuracy on both the random and the ``$\mathrm{T} = \mathrm{298K}$'' test set.
While this approach models can model explicit interactions between the solute and solvent, further modifications of the explainability technique are required to obtain separate relevancy matrices $\mathbf{R}_\mathrm{solu}$ and $\mathbf{R}_\mathrm{solv}$ for the solute and solvent.

\begin{figure}[!h]
\centering
  \begin{subfigure}[t]{0.49\textwidth}
    \centering
    \includegraphics[width=0.99\textwidth]{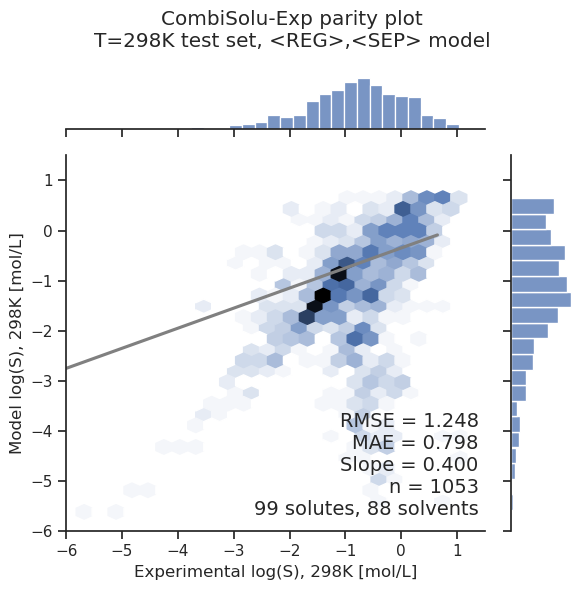}
  \end{subfigure}
  \begin{subfigure}[t]{0.49\textwidth}
    \centering
    \includegraphics[width=0.99\textwidth]{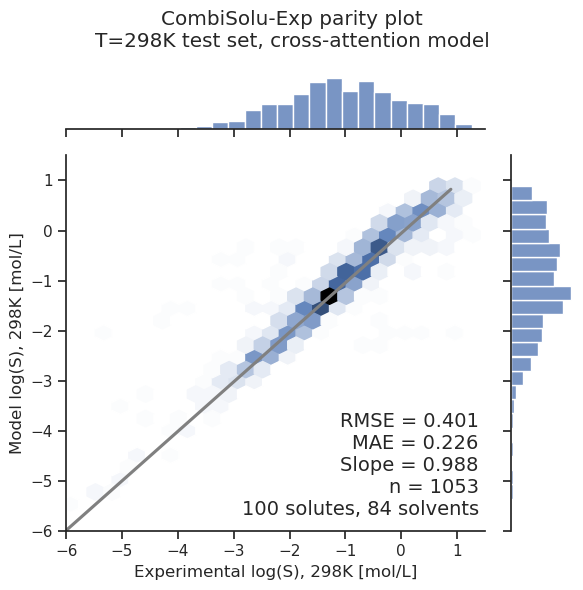}
  \end{subfigure}
\caption{
Parity plots showing the predictive accuracy of the finetuned MegaMolBART model trained on the $\mathrm{T}=\mathrm{298K}$ test set of the CombiSolu-Exp dataset.
The left plot shows the comparatively poor results of the <REG>,<SEP> model variant next to the state-of-the-art results of the cross-attention variant shown on the right.
}
\label{fig:CombiSolv_parityPlots_SI} 
\end{figure}

\newpage

\subsection{AqueousSolu Visualizations.}
Once the relevance weights for each atom token are obtained, the molecules' 2d structure diagram is plotted using the MolDraw2DCairo and DrawMoleculeWithHighlights functions of RDKit. Each atom's attributed weights are scaled using the molecules' minimum and maximum attributed importance to obtain relative importance in arbitrary units and mapped using a color palette before plotting. The light color palette ``green'' of the seaborn library is chosen as the most intuitive.
%%%%%%%%%%%%%%%%%%%%%%

\begin{figure}[!h]
\centering
    \begin{subfigure}[t]{0.41\textwidth}  
    \centering
    \includegraphics[width=\textwidth]{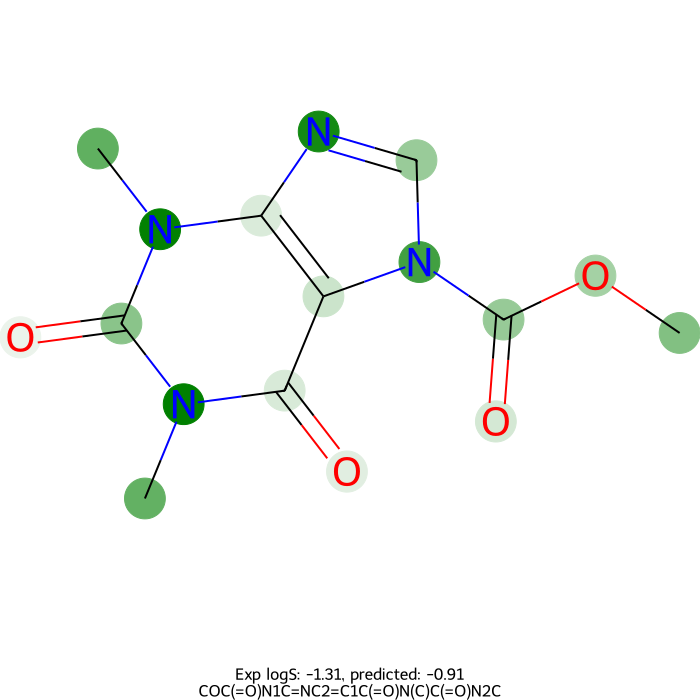}
  \end{subfigure}
  \hfill
    \begin{subfigure}[t]{0.41\textwidth}  
    \centering
    \includegraphics[width=\textwidth]{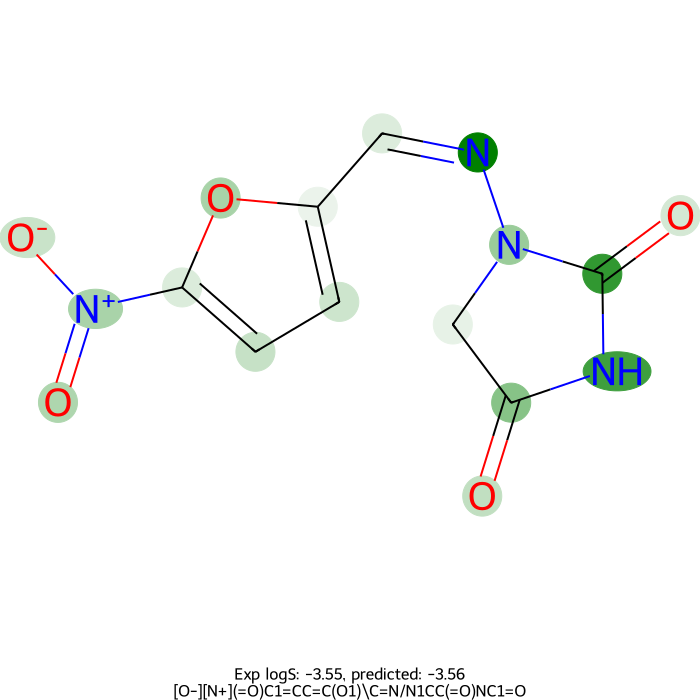}
  \end{subfigure}
  % \hfill
    \begin{subfigure}[t]{0.16\textwidth} 
    \centering
    \includegraphics[width=0.55\textwidth]{figs/colorbar_AU.png}
  \end{subfigure}
  \hfill
  \begin{subfigure}[t]{0.32\textwidth}  
    \centering
    \includegraphics[width=\textwidth]{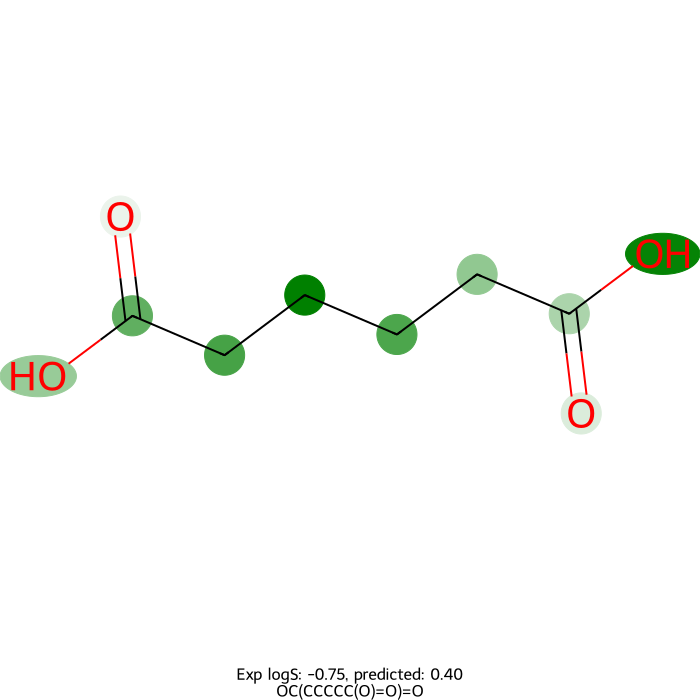}
  \end{subfigure}
  \hfill
  \begin{subfigure}[t]{0.32\textwidth}  
    \centering
    \includegraphics[width=\textwidth]{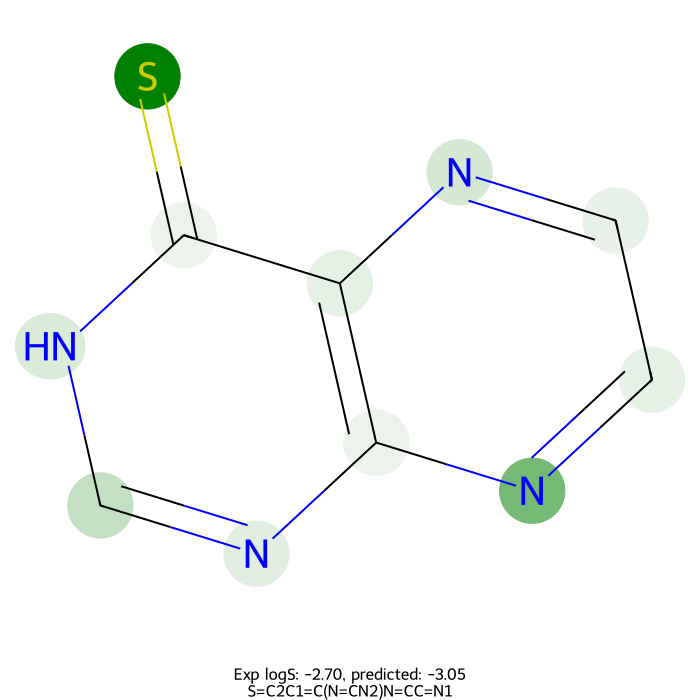}
  \end{subfigure}
  \hfill
    \begin{subfigure}[t]{0.32\textwidth}  
    \centering
    \includegraphics[width=\textwidth]{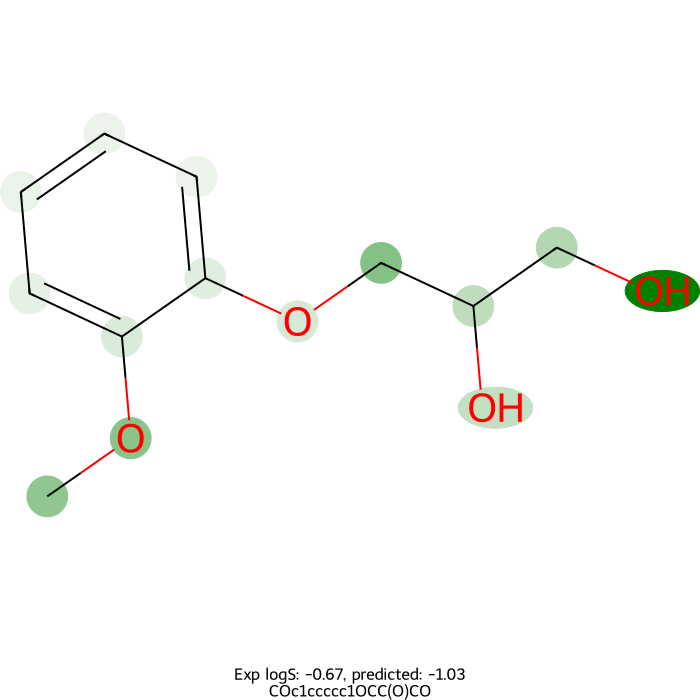}
  \end{subfigure}
  \hfill
  \begin{subfigure}[t]{0.32\textwidth}  
    \centering
    \includegraphics[width=\textwidth]{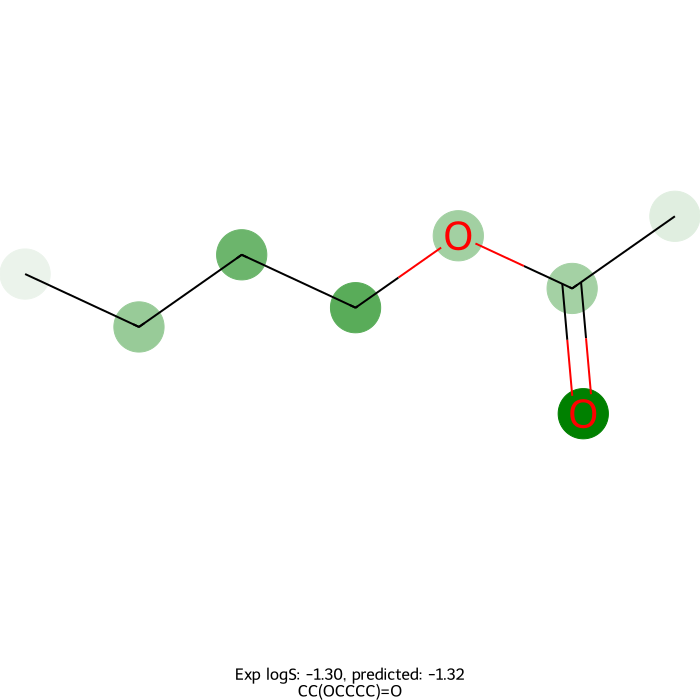}
  \end{subfigure}
  \hfill
  \begin{subfigure}[t]{0.32\textwidth}  
    \centering
    \includegraphics[width=\textwidth]{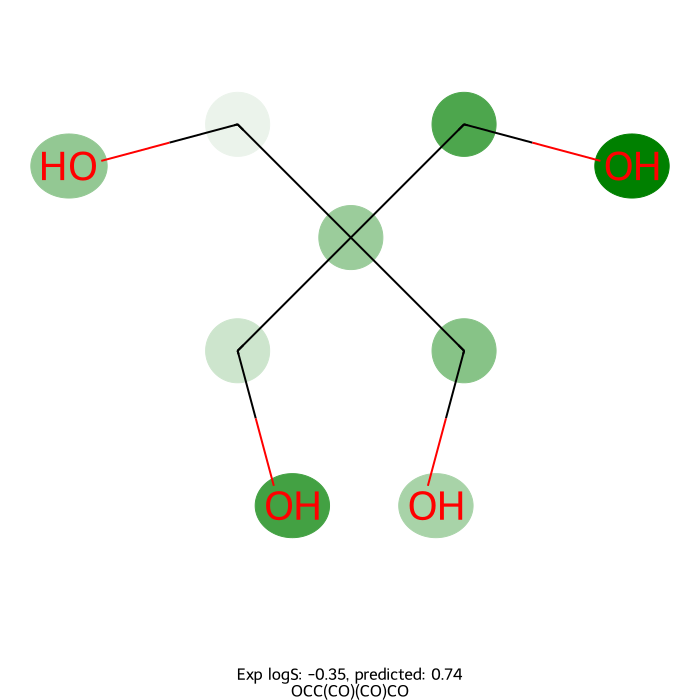}
  \end{subfigure}
  \hfill
    \begin{subfigure}[t]{0.32\textwidth}  
    \centering
    \includegraphics[width=\textwidth]{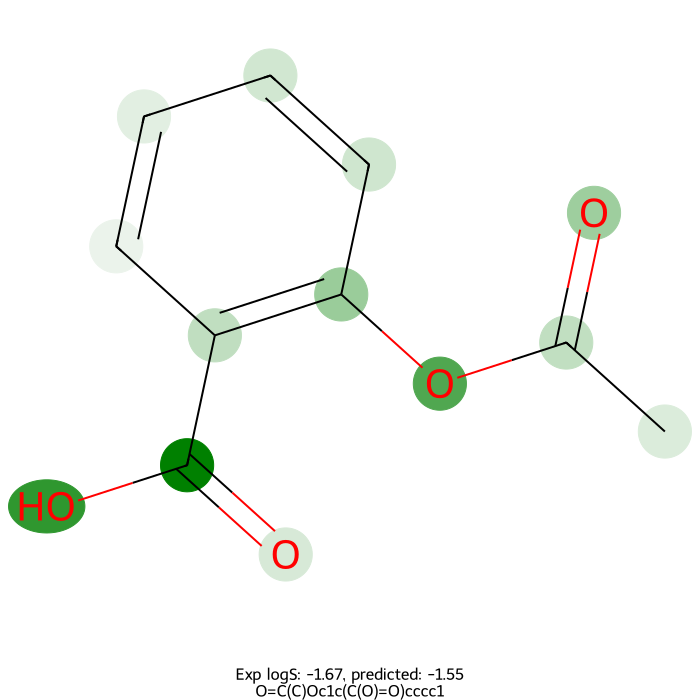}
  \end{subfigure}
  \hfill
  \begin{subfigure}[t]{0.32\textwidth}  
    \centering
    \includegraphics[width=\textwidth]{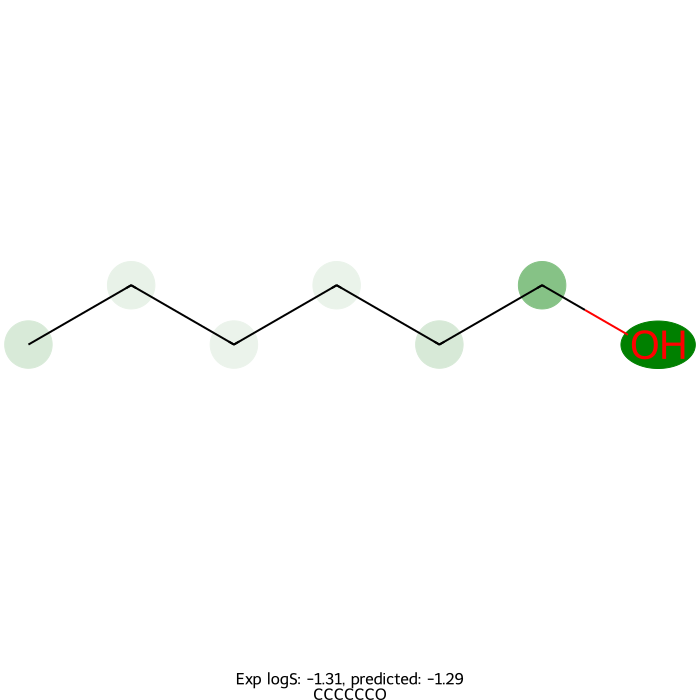}
  \end{subfigure}
  \hfill
  \begin{subfigure}[t]{0.32\textwidth}  
    \centering
   \includegraphics[width=\textwidth]{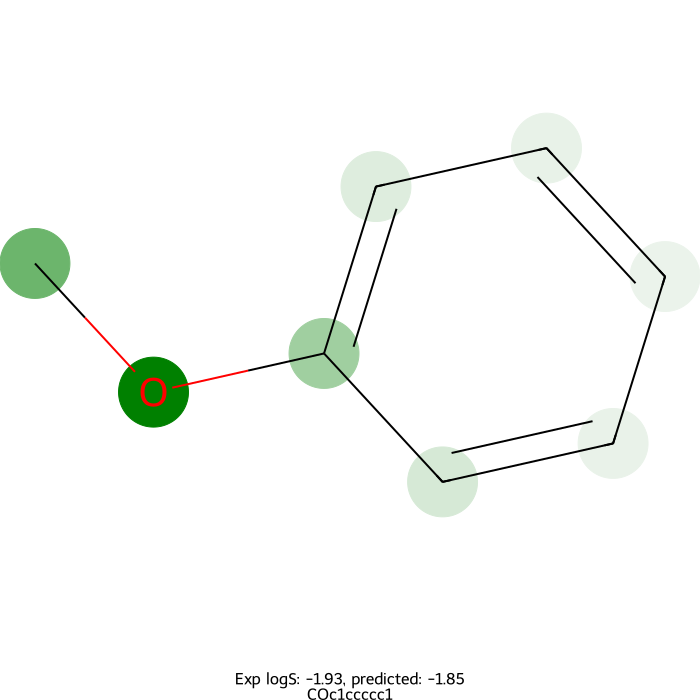}
  \end{subfigure}
  \hfill
    \begin{subfigure}[t]{0.32\textwidth}  
    \centering
    \includegraphics[width=\textwidth]{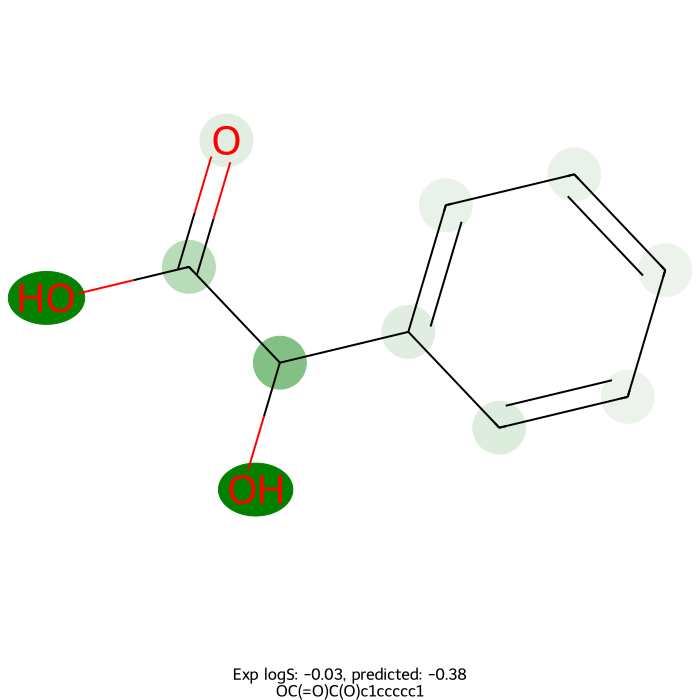}
  \end{subfigure}  
\label{fig:aqueousVisualization_SI} 
\end{figure}

\begin{figure}[!h]
\centering
    \begin{subfigure}[t]{0.32\textwidth}  
    \centering
    \includegraphics[width=\textwidth]{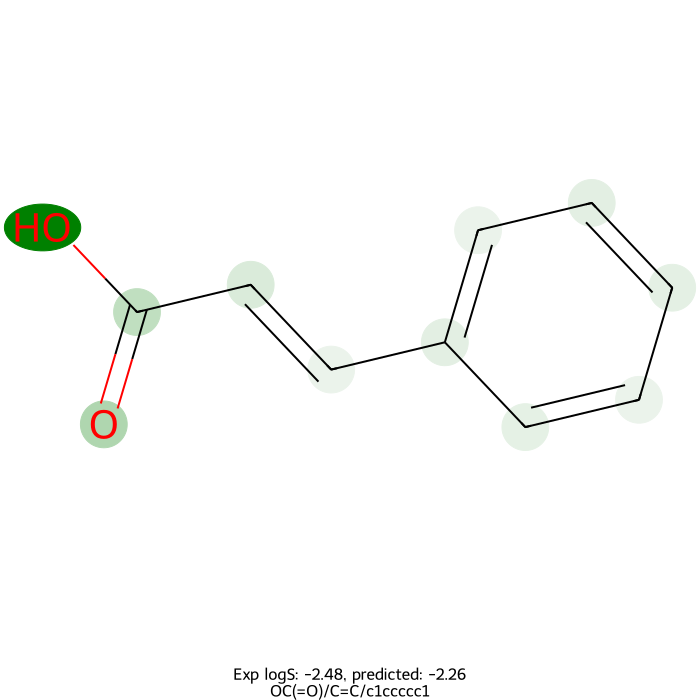}
  \end{subfigure}
  \hfill
  \begin{subfigure}[t]{0.32\textwidth}  
    \centering
     \includegraphics[width=\textwidth]{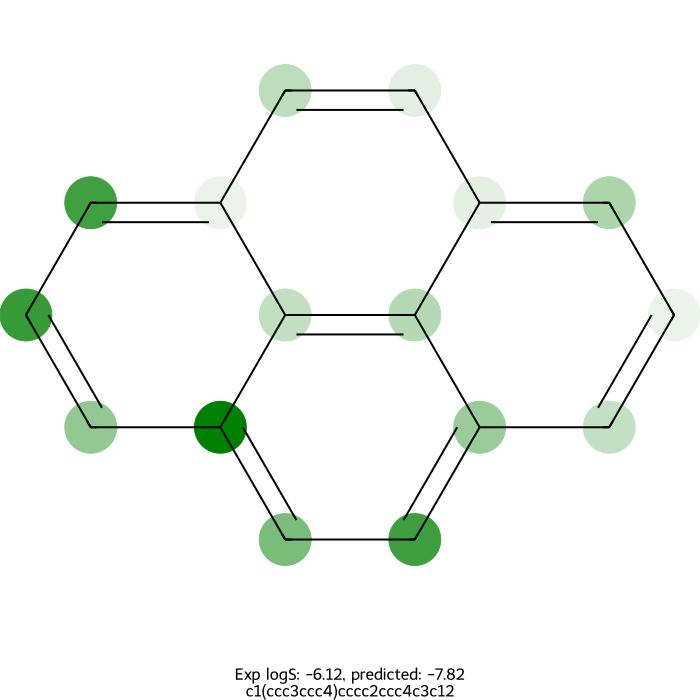}
  \end{subfigure}
  \hfill
    \begin{subfigure}[t]{0.32\textwidth}  
    \centering
    \includegraphics[width=\textwidth]{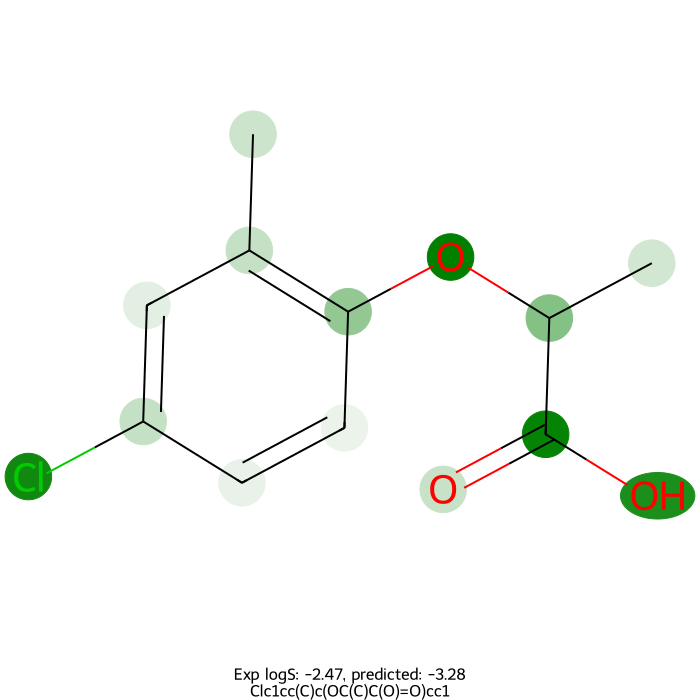}
  \end{subfigure}

\caption{A representative set of visualizations of molecules selected from a random AqueousSolu test set, which showcase what the finetuned MegaMolBART represents. As elaborated in the discussion of results, high importance is generally attributed to the correct functional groups. However, for some molecules relevance is also attributed to the structural carbon backbone, which is likely a result from the pretraining regiment of MegaMolBART. The attributed relevance of symmetric molecules can be poor, but addressing this limitation is not trivial.}
\label{fig:aqueousVisualization_SI_2} 
\end{figure}

\newpage
\subsection{CombiSolu-Exp Visualizations.}

\begin{figure}[!h]
\centering
  \begin{subfigure}[t]{0.49\textwidth}  
    \centering
    \includegraphics[width=\textwidth]{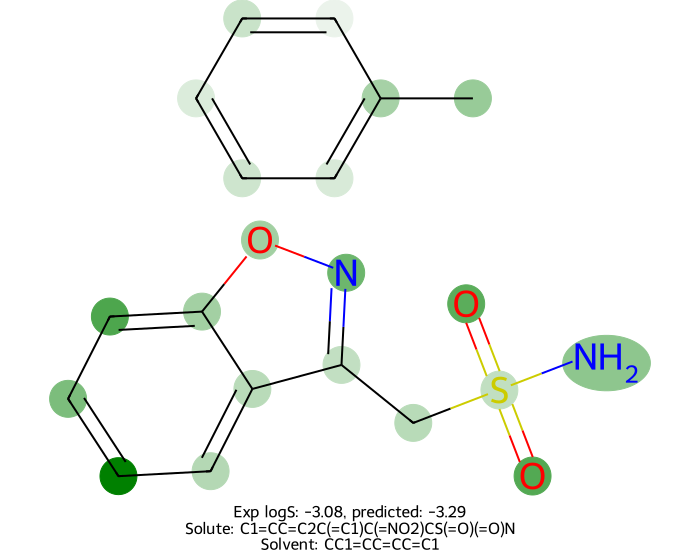}
  \end{subfigure}
  \hfill
  \begin{subfigure}[t]{0.49\textwidth}
    \centering
    \includegraphics[width=\textwidth]{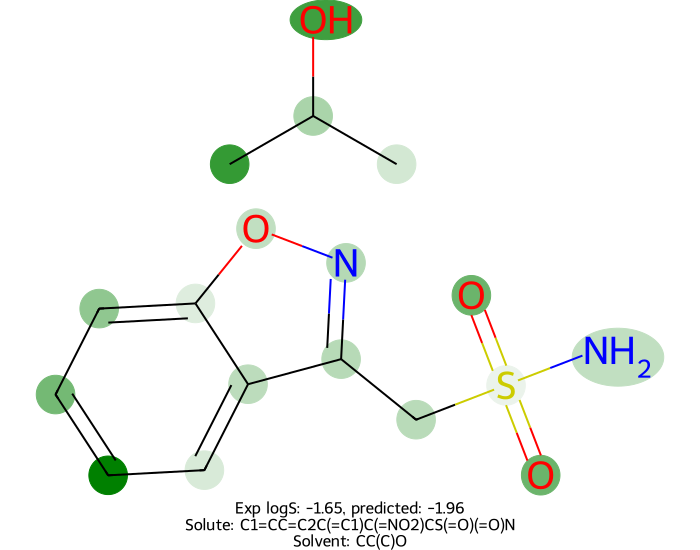}
  \end{subfigure}
  \begin{subfigure}[t]{0.49\textwidth}  
    \centering
    \includegraphics[width=\textwidth]{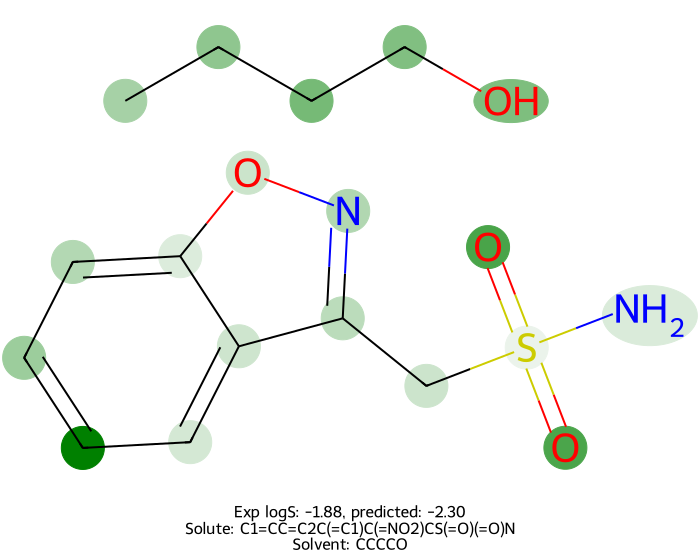}
  \end{subfigure}
  \hfill
  \begin{subfigure}[t]{0.49\textwidth}
    \centering
    \includegraphics[width=\textwidth]{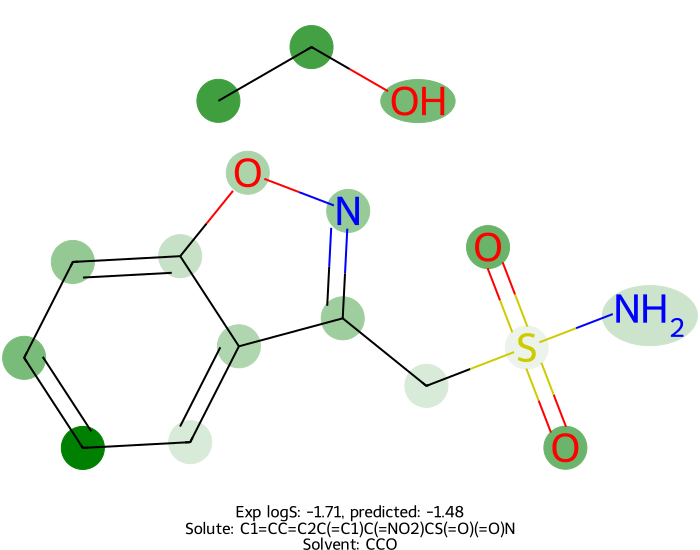}
  \end{subfigure}
\caption{Visualization of attributed relevance of a solute in four solvents, specifically an oil (toluene, top left) and three alcohols (isopropanol, top right; 1-butanol, bottom left; ethanol, bottom right).
The differences in attributed importance of individual atoms or functional groups towards solubility in different solvents can be seen, which are nearly identical for the solute. This is contrary to expectation, which suggests different functional groups should contribute towards solubility in an oil compared to the alcohols.
The visualizations highlight the model does not explicitly model the interaction between solute and solvent. The color bar is the same as in Fig. \ref{fig:aqueousVisualization_SI_2}.}
\label{fig:soluSolvVisualization} 
\end{figure}

\newpage

\newpage
%%%%%%%%%%%%%%%%%%%%%%%%%%%%%%%%%%%%%%%%%%%%%%%%%%%%%%%%%%%%

\end{document}